\documentclass{article}
\usepackage{spconf,amssymb,amsmath,graphicx}
\usepackage{epsfig}
\usepackage{xspace}
\usepackage{srcltx}
\usepackage{caption}
\usepackage{subcaption}
\usepackage[export]{adjustbox}
\usepackage{multirow,makecell,hhline}
\usepackage{booktabs}
\usepackage{textcomp}
\usepackage{cite}
\usepackage{soul}
\usepackage{url}
\usepackage{hyperref}
\hypersetup{colorlinks=True, urlcolor=black}

\usepackage{setspace}

\usepackage{color}
\usepackage[dvipsnames]{xcolor}

\newcommand{\CMT}[1]{{}}

\newcommand{\tbh}[1]{\textbf{#1}}

\title{Efficient Domain Adaptation for Speech Foundation Models}
\name{Bo Li, Dongseong Hwang, Zhouyuan Huo, Junwen Bai, Guru Prakash}
\nameplus{Tara N. Sainath, Khe Chai Sim, Yu Zhang, Wei Han, Trevor Strohman, Francoise Beaufays}
\address{Google LLC, USA \\
\fontsize{9}{9}\selectfont\ttfamily\upshape
\{boboli,tsainath\}@google.com}
\begin{document}
\ninept
\maketitle
\begin{abstract}
Foundation models (FMs), that are trained on broad data at scale and are adaptable to a wide range of downstream tasks, have brought large interest in the research community. Benefiting from the diverse data sources such as different modalities, languages and application domains, foundation models have demonstrated strong generalization and knowledge transfer capabilities. In this paper, we present a pioneering study towards building an efficient solution for FM-based speech recognition systems. We adopt the recently developed self-supervised BEST-RQ for pretraining, and propose the joint finetuning with both source and unsupervised target domain data using JUST Hydra. The FM encoder adapter and decoder are then finetuned to the target domain with a small amount of supervised in-domain data. On a large-scale YouTube and Voice Search task, our method is shown to be both data and model parameter efficient. It achieves the same quality with only 21.6M supervised in-domain data and 130.8M finetuned parameters, compared to the 731.1M model trained from scratch on additional 300M supervised in-domain data.


\end{abstract}
\begin{keywords}
foundation models, domain adaptation
\end{keywords}
\section{Introduction}
\label{sec:intro}

Large models trained on broad data at scale are often desired since they can be adaptable to a wide range of downstream tasks. One prevailing work is the Foundation Model (FM) \cite{bommasani2021opportunities}, which has brought tremendous interest in the research community. FMs have demonstrated strong generalization and knowledge transfer capabilities \cite{brown2020language,adiwardana2020towards}, as a result of learning from diverse data sources such as different modalities, multiple languages and various application domains. In the speech community, there have been numerous research studies showing promising results and demonstrating the potential advantages of such models \cite{NarayananMisraSimPundakEtAl18,chan2021speechstew,chen2021gigaspeech,zhang2022bigssl,radford2022robust,hwang2022pseudo,gandhi2022esb}. Depending on whether the supervised training data is used, we can group the foundation models into two categories, self-supervised pretrained models \cite{devlin2018bert,baevski2020wav2vec} and supervised multitask-trained ones \cite{raffel2020exploring}. With self-supervised training, the models are first trained on audio-only data using contrastive loss \cite{oord2018representation} or reconstruction loss \cite{chorowski2019unsupervised}, to learn good representations of the speech signals. These models are then directly used as feature extractors for downstream tasks \cite{huang2022s3prl,lin2022analyzing}. As no label information is needed, this approach can easily scale up for more diverse speech data without human transcription effort involved. With supervised multitask learning similar to \cite{raffel2020exploring}, different tasks are unified into a heterogeneous discriminative task and the model is trained jointly on these tasks, such as multi-domain tasks \cite{narayanan2018toward,chan2021speechstew} or multilingual tasks \cite{li2021scaling, li2022massively}. A prerequisite of this approach is to have some labeled data for tasks that the FMs are trained on. Recent works also found that self-supervised pretraining could improve the label data efficiency for the supervised multitask models \cite{chan2021speechstew,zhang2022bigssl}. In this paper, we mainly focus on FMs trained under this procedure for better recognition quality. 

Existing work has mainly focused on using supervised in-domain data to jointly train or finetune FMs for target tasks \cite{chan2021speechstew,zhang2022bigssl}. The use of supervised in-domain data requires FM retraining when new domain presents. To address this, we propose to build FMs from the public domain YouTube data. YouTube has a diverse source of speech from more than 100 different countries around the world, across 80 languages and covers a large variety of domains \cite{liao2013large}. The amount of data is also tremendous \cite{youtubestats}. These make it a great source for building speech FMs, which can also be shared between industry and academia to foster collaborations. For better quality, FMs tend to have large sizes. Fine tuning such models are resource inefficient and time consuming. Techniques that can efficiently adapt FMs to a target task are crucial. Existing work such as residual adapters \cite{houlsby2019parameter,hwang2022large,biadsy2022scalable}, prompting \cite{he2021towards} and neural reprogramming \cite{yang2021voice2series} have demonstrated such potentials. 

Our work contributes to FM learning in several aspects. First, conventional pretrain+finetune 2-stage scheme like wav2vec 2.0 \cite{baevski2020wav2vec} and wav2vec-BERT \cite{chung2021w2v} updates the pretrained encoder during finetuning, which, however, is often costly for large FMs. Therefore, we propose to freeze the pretrained FMs during finetuning and only update a cheap added adapter for the target domain. We show such method works for CTC, LAS, and RNN-T decoders.
Second, most FMs finetune the model with labeled data from the target domain. We demonstrate proper data selection from the source domain (YouTube) can also mitigate the gap between the source and target domain, and further help the performance in the target domain (Voice Search).
Third, even without audio-text paired data from the target domain, we show either audio-only or text-only data can facilitate the FM finetuning.
We present a full recipe of building a FM that can be efficiently adapted to a target domain. Despite some techniques in this paper have been studied previously, this paper innovatively introduces a unified framework for building high quality speech solutions for downstream tasks using FMs. In this framework, we explore data efficiency and model efficiency techniques to build FMs that have a coarse connection to the target domain but can be fast adapted with limited amount of supervised in-domain data. This opens new research directions to explore techniques to push the limit of FMs for speech tasks.







\section{Methods}
\label{sec:techs}

This study focuses on domain adaptation of FMs built on YouTube data, to the target domain Voice Search. We strive for cost-efficient adaptation to reuse the large FMs and minimize the effort in finetuning. Besides, the pressing privacy concerns urge us to better protect the personal data like Voice Search from the extensive ASR training, and more effectively utilize the public data like YouTube. In this section, we propose several efficient adaptation techniques for better ASR quality. 

\subsection{Speech Foundation Model}

We adopt the 600M ConformerXL \cite{zhang2022bigssl} architecture for building our speech foundation model. The source domain, YouTube, contains an extremely wide range of sub-domains \cite{narayanan2018toward,liao2013large}. Such variety is necessary to ensure a good generalization capability of the FM and will be beneficial to various downstream tasks. 
In the first stage, the FM is pretrained on audio-only data. Different from BigSSL \cite{zhang2022bigssl}, BERT-based Speech pre-Training with Random-projection Quantizer (BEST-RQ) is adopted as it outperforms wav2vec 2.0 and wav2vec-BERT on many tasks \cite{chiu2022self}. This technique randomly initializes a projection matrix and a codebook. Each input speech frame is projected into a hidden space with this random matrix. The index of the nearest vector from the random codebook is used as the discretized label token for this frame. We can then apply BERT training on this discretized representation of the speech signal, which is firstly masked, and then the model is optimized to predict the label tokens of the masked part. During training, both the randomly initialized projection matrix and the codebook are kept frozen. The input speech data is normalized to have zero mean and unit standard deviation. We find this normalization is critical for preventing the random projection from collapsing to a small subset of the codebook. 
In the second stage, we conduct supervised finetuning on the source domain data. This finetuned model is used as the baseline FM.

\subsection{Domain Adaptation}

Domain adaptation connects the speech FM with a particular application scenario. Given the large scale model size and diverse training data of the FM, efficient adaptation is the key to enable the practical adoption for various downstream tasks. In this work, we optimize for two goals:
\begin{itemize}
    \item \textbf{model efficiency}: It is preferable to train only a small amount of parameters on the target domain and maximize the amount of model parameters that can be shared across domains and tasks. {\it The less the amount of parameters fine-tuned on the in-domain data the better}.
    \item \textbf{data efficiency}: The self-supervised training on diverse data and the large model capacity enables the FM to have better label data efficiency on downstream tasks \cite{zhang2022bigssl}. Besides, it would be even better if we could utilize unparalleled in-domain data such as speech-only and text-only data. {\it The less supervised data and the less in-domain data needed, the better.}
\end{itemize}
Quality-wise, the adaptation technique should reach similar quality as finetuning the whole model with a large amount of supervised in-domain data.

\vspace{-0.1in}
\subsubsection{Model efficiency}

Given the quality benefit of supervised FMs, it would be interesting to understand whether different decoders affect the quality in the target domain. Among the three commonly used decoders CTC \cite{Graves06}, LAS \cite{Chan15} and RNN-T \cite{Graves12}, CTC is the simplest one which basically adds only a single softmax layer on top of the encoder. LAS uses attention to summarize the whole utterance for each decoding step which can have potential quality benefits, while others typically do not explicitly utilize future frames. Moreover, RNN-T has shown better long-form robustness than attention models \cite{narayanan2019recognizing}. We hence compare all these three decoders in this paper. 

On the encoder side, we adopt the residual adapters \cite{houlsby2019parameter} to keep the FM encoder frozen while adapting to the target domain. In the literature there are also many other variants such as prefix tuning, LoRA and Parallel Adapters \cite{he2021towards}. However, in our tasks we did not see quality gains of those variants and only employ the vanilla residual adapter.

\vspace{-0.11in}
\subsubsection{Data efficiency}

The goal of data efficiency is to reduce the amount of in-domain data used for adaptation. The less supervised in-domain speech-text pair the better. Moreover, using unpaired audio or text data is more preferable to the paired data. 

\textbf{Source domain data filtering:} The YouTube source domain contains largely diverse sub-domains such as different topics, styles, speakers, etc. It is likely there are data resembling the target domain. To improve the FM quality on target domain we can filter out such data and customize the FM with them. This can potentially reduce the amount of in-domain data needed. For Voice Search, the speech data is mostly single speaker and for search purpose, written form transcripts are normally preferred. Based on these assumptions, we filter YouTube data to find utterances that only contain a single speaker via speaker clustering, and the written and spoken forms of the transcripts are the same. We find this help reduce FM's recognition errors on the target domain without any in-domain data. There are many other filtering techniques such as \cite{park2022unsupervised}, which will be explored in the future. 

\textbf{Audio only in-domain data:} One potential mismatch between the source and target domains is the acoustic difference. To improve the quality of the FM on the target domain, we can bring in the audio only in-domain data to address this mismatch. Joint supervised and unsupervised training (JUST) \cite{bai2022joint} and its variant JUST Hydra \cite{hwang2022pseudo} have demonstrated the advantage of joint training. In this setup, we combine the source domain paired speech-text data (supervised) with the target domain audio-only data (unsupervised) via joint training. Alternatively, noisy student training (NST) \cite{park2020improved,xie2020self} is another popular approach to utilize unsupervised speech with a teacher model by providing pseudo-labels for our training. We find it very effective in improving the quality.

\textbf{Text only in-domain data:} Another possible mismatch comes from the language model differences between domains. The words used and how they are used can vary across domains. To address that, we can inject target domain text data into the FM. The simplest approach is to rescore the recognition hypotheses with an LM trained in the target domain. MaxEnt LM has shown very promising results for Voice Search tasks \cite{biadsy2017effectively}. Besides rescoring, fusing the target domain LM in the beam search decoding has also shown better qualities \cite{toshniwal2018comparison}, which is covered in this work. More recently, injecting text data via joint training such as SLAM \cite{bapna2021slam}, MAESTRO \cite{chen2022maestro} and JOIST \cite{sainath2022joist} have shown promising gains, which will be left for future work.

\section{Experimental Setup}
\label{sec:exp}

\subsection{Data}

\subsubsection{YouTube}

For the source domain, we collected two datasets based from YouTube. The unsupervised YouTube data, namely \textbf{YT-U}, is a multilingual YouTube dataset segmented using voice activity detection models \cite{purwins2019deep}. This set brings a diverse speech variations for the FM pretraining.
The supervised YouTube data, namely \textbf{YT-T}, is an English only dataset  from videos that have user-uploaded transcripts. These videos are first segmented using a 100M-parameter RNN-T model with a bi-directional LSTM encoder \cite{chiu2021rnn}. The non-speech segments identified by the YT teacher model are removed to yield approximately 500k hours of unlabeled audio data. The user provided transcripts, however, are discarded and we generate pseudo-labels using the same YT teacher model. 
The test set for the source domain is generated by hand transcribing popular videos from YouTube with 11 hrs of audio with lengths 2 - 10 min. 

\subsubsection{Voice Search}

Similarly, for the target domain, we use two datasets collected from the English Voice Search traffic. The unsupervised Voice Search data, namely \textbf{VS-U}, contains 383.7M utterances which corresponds to 532.9k hours of speech data. The data is deidentified and not transcribed. It is either used without transcripts or with teacher model generated machine transcripts. 
The supervised Voice Search data, namely \textbf{VS-L}, contains totally 21.6M English utterances from different English speaking countries. It corresponds to 30k hours of Voice Search traffic data, which is deidentified and human transcribed. 
The unpaired text data, namely \textbf{VS-Text} consists of more than 100B utterances and is thus much larger than our audio sets. In addition, we incorporate all text data from the VS-L set.
The test set contains 10k utterances sampled from English Voice Search traffic with no overlapping with the training set. They are human transcribed. 
All data are deidentified and the collection and handling abide by Google AI Principles \cite{googleaiprinciples}.

\subsection{Model Architecture}

The input log Mel filterbank features to the network are first passed through a convolutional sub-sampling module to change the 10ms frame rate to 40ms. After that, Conformer \cite{gulati2020conformer}, the convolution-augmented transformer, is used. Each Conformer layer consists of 8-head attention, feed-forward and convolutional modules. There are totally 24 Conformer blocks with a model dim of 1024. An output projection layer is used after the Conformer blocks to generate the final encoder outputs.
These encoder outputs are either used as input for a CTC \cite{Graves06} model after an additional projection layer or passed to a RNN-T \cite{Graves12} decoder or LAS \cite{Williams18} decoder along with 6 LSTM layers. Each LSTM layer has a cell dim of 768 and a hidden dim of 3072 following \cite{zhang2022bigssl}.
128-dim residual adapters are used when adapting the FM towards the target Voice Search domain.

\subsection{Model Training}

The speech data used in this work are uniformly sampled to 16 KHZ quality. Any data with a different native sampling rate is either up-sampled or down-sampled. We use 128-dim log Mel features that are computed using 32ms windows with a 10ms hop. SpecAugment \cite{Park_2019} is used to improve models' robustness against noise. Specifically, two frequency masks with a maximum length of 27 and two time masks with a maximum length of 50 are used. 4K word piece model (WPM) is used to tokenize the training transcripts.

All the models are trained in Tensorflow using the Lingvo \cite{shen2019lingvo} toolkit on Google's Tensor Processing Units (TPU) V3 \cite{tpu} with a global batch size of 4,096 utterances. Models are trained with 512 TPU cores and optimized using synchronized stochastic gradient descent. Adafactor \cite{shazeer2018adafactor} with parameters $\beta_1$=0.9 and $\beta_2$=0.99 is used. A transformer learning rate schedule \cite{Vaswani17} with peak learning rate 3e-4 and 10K warm-up steps is used. Exponential moving average is used to stabilize the model weight updates.

\vspace{-0.1in}
\section{Results}
\label{sec:results}
\vspace{-0.05in}

In this section we present our experimental study on adapting a foundation model trained on YouTube to the target Voice Search domain. To justify the effectiveness, we took an existing Conformer RNN-T model with the same architecture but trained on  VS-L together with another 300M English multidomain audio-text pairs\cite{NarayananMisraSimPundakEtAl18}. This model has a WER of 4.3\% on the VS test set. This is referred to as the target domain model ``T0''. The goal of this study is trying to reach that target with less supervised in-domain data (i.e. only the 21.6M VS-L) and also less amount of adaptable parameters. 

\subsection{Foundation Models}
\label{sec:fm}

We pretrain the FM encoder use BEST-RQ based self-supervised training on YT-U (S0 in Table~\ref{tbl:sfm}). We then add an LAS decoder following \cite{zhang2022bigssl}, and finetune only the decoder on YT-L (S1). The high WER suggests the self-supervised trained encoder is not well optimized for the ASR task and it is not suitable to be directly used as an ASR encoder. Next, we finetune the whole model (S2). A 12.4\% WER on VS is obtained. Due to the long form problem of attention based model \cite{chiu2019comparison}, we replace the LAS decoder with a RNN-T decoder (S3), which has a slightly better WER on the short form VS test set, but much better results on the long form YT test set. We tried to use less number of decoder layers (S4) and also a CTC decoder without any layers between the encoder and the output softmax layer (S5). Both of them does worse than S3 on VS. We hence take S3 as our baseline FM for the rest of the study, unless otherwise indicated.

\begin{table}[!ht]
\caption{WER (\%) qualities of different foundation models pretrained (PT) on YT-U and finetuned (FT) on YT-T.}
\centering
\vspace{-0.1in}
\scalebox{0.775}{
\begin{tabular}{lccrc|rr}
\toprule 
\multirow{2}{*}{\tbh{ID}} & \multicolumn{3}{c}{\tbh{Decoder}} & \multirow{2}{*}{\tbh{Train}} & \multirow{2}{*}{\tbh{YT}} & \multirow{2}{*}{\tbh{VS}} \\
\cmidrule{2-4}
~ & \tbh{\scriptsize{Type}} & \tbh{\scriptsize{\# Layers}} & \tbh{\scriptsize{\# Params (M)}} & ~ & ~ & \\
\midrule
\midrule
S0 & - & - & - & PT & - & - \\
S1 & LAS & 6 & 164.3 & FT dec & 92.2 & 65.7 \\
\midrule
S2 & LAS & 6 & 164.3 & FT all & 62.8 & 12.4 \\
S3 & RNN-T & 6 & 124.4 & FT all & 14.3 & 12.3 \\
S4 & RNN-T & 2 & 39.4 & FT all & 14.0 & 12.5 \\
S5 & CTC & 0 & 2.6 & FT all & 13.6 & 13.9 \\
\bottomrule
\end{tabular}
}
\label{tbl:sfm}
\end{table}

\subsection{Data efficient adaptation}
\label{sec:filter}

Despite the large data diversity in YouTube, S3's 12.3\% is far from the in-domain model T0. To improve the quality on the target domain, we can mine data from the source domain that matches with the target domain. For quick iteration, we use CTC decoder (i.e. S5) which uses less memory and is much faster to train, for data selection validation. We vary the speaker similarity threshold to generate different subsets of the YT-L data and then report the WERs of finetuning S6 on each corresponding subset in Table~\ref{tbl:filter}. This simple heuristic filtering reduces the WER on target domain and also reduces the amount of data needed during the fine tuning stage. 

\begin{table}[!ht]
\caption{WER (\%) comparisons of using filtered YT-T data.}
\centering
\vspace{-0.1in}
\scalebox{0.8}{
\begin{tabular}{lr|rr}
\toprule 
\tbh{ID} & \tbh{YT-T Hours (K )} & \tbh{YT} & \tbh{VS} \\
\midrule
\midrule
S5 & 500.0 & 13.6 & 13.9 \\
\midrule
D1 & 456.7 & 13.8 & 13.0  \\ 
D2 & 271.8 & 13.9 & 13.2 \\ 
D3 & 154.9 & 13.9 & 13.2 \\ 
D4 & 13.4 & 13.9 & 14.2 \\ 
\bottomrule
\end{tabular}
}
\label{tbl:filter}
\end{table}

All the existing experiments used only the source domain data, though the quality is decent on the target domain, it is still far from the in-domain model T0. 
To address this, we started with unsupervised speech data and adopted the JUST Hydra \cite{hwang2022pseudo} training. It combines the supervised fine tuning loss on YouTube data with the self-supervised loss on the Voice Search audio-only data. Adding joint training to S3 (G1), we can reduce VS WER from 12.3\% to 6.9\%, which clearly demonstrate the benefit of in-domain data. 


Another popular approach of using audio only data is to do noisy student training (NST) \cite{xie2020self}. 
We first take model S3 that is trained only on the YouTube domain as the teacher model (G2). The use of teacher transcriptions does help reduce the WER from 12.3\% to 11.6\%, however the high teacher WER limits the improvement. Instead, when we use the target-domain model T0 as teacher (G3), we can achieve a 4.5\% WER even with just 10\% of the VS-U data. This confirms the necessity of building high quality in-domain teachers and also suggests there's large redundancy in the VS-U set. Developing effective data selection techniques would be useful and will be explored in future work.

\begin{table}[!ht]
\caption{WER (\%) comparisons of using unpaired target domain data (VS-U, VS-Text).}
\centering
\vspace{-0.1in}
\scalebox{0.8}{
\begin{tabular}{lcr|r}
\toprule 
\tbh{ID} & \tbh{Data Type} & \tbh{Data Ratio} & \tbh{VS} \\
\midrule
\midrule
S3 & - & - & 12.3 \\
\midrule
G1 \scriptsize{JUST Hydra} & VS-U & 100\% & 6.9 \\
G2 \scriptsize{NST with S3} & VS-U & 100\% & 11.6 \\
\midrule
G3 \scriptsize{NST with T0} & VS-U & 100\% & 4.5 \\
 & VS-U & 10\% & 4.5 \\
 & VS-U & 3\% & 4.6 \\
 & VS-U & 1\% & 5.1 \\
\midrule
H1 \scriptsize{Rescoring} & VS-Text & 100\% & 11.1 \\ 
H2 \scriptsize{Shallow Fusion} & VS-Text & 100\% & 12.0 \\
\bottomrule
\end{tabular}
}
\label{tbl:audio}
\end{table}

Text is another source that can be easily obtained for the target domain than paired speech-text data. We investigated two basic techniques, namely MaxEnt LM rescoring (H1 in Table~\ref{tbl:audio}) \cite{biadsy2017effectively} and Conformer LM Shallow Fusion (H2)\cite{Le2021}, to inject VS-Text into the YouTube only FM. Both approach reduces WER on VS but the gain is relatively small. Many errors are due to transcribing background speech caused by the YouTube and Voice Search domain mismatch. This suggests later text injection may be preferred. 

From these ablation studies, we can improve the FM quality on the target domain without any supervised target domain data by filtering source domain data, joint training with target domain audio only data and integration with target domain LMs. However there's still quality gap from the target domain model T0.


\subsection{Parameter efficient adaptation}
\label{sec:param}


Given training with in-domain data is the most effective approach, we look into efficient model adaptation with in-domain data. We adapt the YT FM to VS using a small amount of supervised in-domain data VS-L (Table~\ref{tbl:adapt}). We took S3 and finetuned the whole model on VS-L (E1). It yields a 4.4\% WER on VS similar to T0's 4.3\%, demonstrating the use of pretrained FM improves the label data efficiency \cite{zhang2022bigssl}. 

The encoder has 606.6M parameters, around 83\% of the the whole model, we thus want to freeze this part to limit the amount of finetuned parameters. Only adapting the decoder (E2) gives a 8.1\% WER. With residual adapters inserted between each encoder layer (E3), we can obtained a 6.7\% WER with only 6.4M parameters adapted. When combining E2 and E3, we can achieve a WER of 4.5\% with only 130.8M parameters (E4). E4 is more parameter efficient than E1 (130.8M vs. 731.1M adaptable parameters) and E4 is also more data efficient than T0 (around 20M vs. 320M supervised data) to achieve similar quality.

\begin{table}[!t]
\caption{WER (\%) comparisons of finetuning using VS-L data.}
\centering
\vspace{-0.1in}
\scalebox{0.8}{
\begin{tabular}{lcrr|r}
\toprule 
\multirow{2}{*}{\tbh{ID}} & \multirow{2}{*}{\tbh{Comp.}} & \multirow{2}{*}{\tbh{\# Params}} & \multirow{2}{*}{\tbh{Training Speed}} & \multirow{2}{*}{\tbh{VS}} \\
~ & ~ & \tbh{\scriptsize{(M)}} & \tbh{\scriptsize{(utterances/second)}} \\
\midrule
\midrule
S3 & - & - & - & 12.3 \\
\midrule
E1 & full model & 731.1 & 3100 & 4.4 \\
E2 & dec. & 124.4 & 3600 & 8.1 \\
E3 & enc. adapter & 6.4 & 4989 & 6.7 \\
E4 & enc. adapter + dec. & 130.8 & 3500 & 4.5 \\
\bottomrule
\end{tabular}}
\label{tbl:adapt}
\vspace{-0.2in}
\end{table}

\subsection{Final Recipe}


Finally, we combine the various techniques explored in previous sections to form the final recipe for our FM and target domain adaptation. It consists: 
\begin{enumerate}
    \item F1: BEST-RQ based self-supervised training on YT-U (S0); 
    \item F2: Adding a RNN-T decoder with 6 LSTM layers (S3) and conducting JUST Hydra training with filtered YT-T (D1) and VS-U data (G1);
    \item F3: Inserting residual adapters between each conformer layers (E3) and finetuning encoder adapter and decoder on target domain supervised data VS-L (E4).
\end{enumerate}
We didn't see quality gains with H1 or H2 on F3. With these, we obtain a similar 4.4\% WER on VS to the in-domain model T0, but with much less in-domain supervised data (21.6M vs. 321.6M) and much less finetuned parameters (130.8M vs. 731.1M).

\begin{table}[!ht]
\caption{WER (\%) qualities of different stages in the final recipe.}
\centering
\vspace{-0.1in}
\scalebox{0.95}{
\begin{tabular}{ll|r}
\toprule 
\tbh{ID} & \tbh{Data} & \tbh{VS} \\
\midrule
\midrule
F1 \scriptsize{S0, pretrain} & YT-U & - \\
F2 \scriptsize{6L RNN-T decoder, JUST} & YT-T(D1), VS-U & 6.9 \\
F3 \scriptsize{enc. adapter + dec.} & VS-L & 4.4 \\
\bottomrule
\end{tabular}
}
\label{tbl:final}
\end{table}
\vspace{-0.1in}
\section{Conclusions}
\label{sec:concl}
\vspace{-0.05in}

In this paper, we investigate the problem of efficient domain adaptation for speech foundation models (FMs). On a task that uses YouTube as the source domain where FMs are trained on and Voice Search as the target domain, we empirically studied different techniques to improve the quality of FMs on the target domain. To achieve data efficiency, we demonstrated the potential of using speech-only and text-only data to improve the quality of FM on the target domain. However, there are still large gap from the model trained with in-domain paired speech-text data. To achieve parameter efficiency, we  adapt the FM with encoder adapter and decoder to reach the same quality of full model finetuning. The techniques investigated in this paper are just a small subset of methods that can be used for efficient domain adaptation for speech FMs. There are more questions to answer and potential directions to explore. We hope this study would bring more interest to the paradigm of foundation model based speech solutions.

\vfill\pagebreak


\begin{spacing}{0.9}
\bibliographystyle{IEEEbib}
{\footnotesize\bibliography{main}}
\end{spacing}

\end{document}